\documentclass{article}
\usepackage[T1]{fontenc}
\usepackage[utf8]{inputenc}
\usepackage{amsmath}
\usepackage{graphicx}
\usepackage{microtype}
\usepackage[unicode=true,
 bookmarks=false,
 breaklinks=false,pdfborder={0 0 1},backref=section,colorlinks=false]
 {hyperref}

\makeatletter

\usepackage[final,nonatbib]{nips_2018}

\usepackage{url}
\usepackage{amsfonts}
\usepackage{nicefrac}
\usepackage[english]{babel}

\DeclareMathOperator*{\argmax}{arg\,max}

\title{Chess2vec: Learning Vector Representations for Chess}

\author{
  Berk Kapicioglu\\
  OccamzRazor\\
  \texttt{berk@occamzrazor.com}
  \And
  Ramiz Iqbal\thanks{Research conducted while author was an intern at OccamzRazor.}\\
  MD Anderson Cancer Center\\
  \texttt{riqbal@mdanderson.org}
  \And
  Tarik Koc\\
  OccamzRazor\\
  \texttt{tarik@occamzrazor.com}
  \And
  Louis Nicolas Andre\\
  OccamzRazor\\
  \texttt{louis@occamzrazor.com}
  \And
  Katharina Sophia Volz\\
  OccamzRazor\\
  \texttt{volz@occamzrazor.com}
}

\makeatother

\begin{document}
\maketitle
\begin{abstract}
We conduct the first study of its kind to generate and evaluate vector
representations for chess pieces. In particular, we uncover the latent
structure of chess pieces and moves, as well as predict chess moves
from chess positions. We share preliminary results which anticipate
our ongoing work on a neural network architecture that learns these
embeddings directly from supervised feedback. 
\end{abstract}
The fundamental challenge for machine learning based chess programs
is to learn the mapping between chess positions and optimal moves
\cite{Lai2015,David2016,Silver2017}. A chess position is a description
of where pieces are located on the chessboard. In learning, chess
positions are typically represented as bitboard representations \cite{wiki:bitboard}. 

A bitboard is a $8\times8$ binary matrix, same dimensions as the
chessboard, and each bitboard is associated with a particular piece
type (e.g.\@ pawn) and player color (e.g.\@ black). On the bitboard,
an entry is $1$ if a piece associated with the bitboard's type and
color exists on the corresponding chessboard location, and $0$ otherwise.
Each chess position is represented using $12$ bitboards, where each
bitboard corresponds to a different piece type (i.e.\@ king, queen,
rook, bishop, knight, or pawn) and player color (i.e.\@ white or
black). More formally, each chess position is represented as a $8\times8\times12$
binary tensor. 

The bitboard representation can alternatively be specified by assigning
a $12$-dimensional binary vector to each piece type and color. Under
this interpretation, given a chess position, each of the $64$ chessboard
locations is assigned a $12$-dimensional vector based on the occupying
piece. If the location has no piece, it is assigned the zero vector,
and if it has a piece, it is assigned the sparse indicator vector
corresponding to the piece type and color. Note that, with this representation,
all black pawns would be represented using the same vector. 

In this paper, our goal is to explore new $d$-dimensional vector
representations for chess pieces to serve as alternatives to the $12$-dimensional
binary bitboard representation. 

\section{Latent structure of pieces and moves \label{sec:Latent-structure-of-chess}}

In this section, we analyze the latent structure of chess pieces and
moves using principal components analysis. We conduct the analysis
on two different datasets: one which contains legal moves and another
which contains expert moves. Aside from gaining qualitative insights
about chess, we also obtain our first vector representations. 

\subsection{Principal component analysis (PCA)}

PCA is a method for linearly transforming data into a lower-dimensional
representation \cite{Gareth2013}. In particular, let data be represented
as the matrix $X\in\mathbb{R}^{n\times p}$ whose columns have zero
mean and unit variance. Each row of $X$ can be interpreted as a data
point with $p$ features. Then, PCA computes the transformation matrix
$C^{\star}\in\mathbb{R}^{p\times d}$ by solving the optimization
problem
\begin{equation}
\begin{aligned} & \underset{C}{\text{minimize}} &  & \left\Vert X-XCC^{T}\right\Vert _{F}^{2}\\
 & \text{subject to} &  & C^{T}C=I
\end{aligned}
.\label{eq:pca-objective}
\end{equation}
Intuitively, $C^{\star}$ linearly transforms the data from a $p$-dimensional
space to a $d$-dimensional space in such a way that the inverse transformation
(i.e.\@ reconstruction) minimizes the Frobenius norm. The mapping
of $X$ into the $d$-dimensional space is given by $W=XC^{\star}$.
The importance of minimizing reconstruction error will be clear when
we quantitatively evaluate the embeddings obtained via PCA.

In order to apply PCA to chess, we need an appropriate way to represent
chess data as $X\in\mathbb{R}^{n\times p}$. Ideally, we would like
pieces with similar movement patterns to have similar low-dimensional
representations, so we represent each piece by its move counts. In
its simplest form, a chess move is specified via a source board location
that a piece is picked from and a target board location that the piece
is moved into. This formulation allows us to avoid specifying the
type of the moving piece or whether the target location already contained
another piece. There are $64$ board locations, thus all move counts
can be represented via a $4096$-dimensional vector, and since there
are $6$ piece types, we let $X\in\mathbb{R}^{6\times4096}$.

\subsection{Stockfish moves}

We generate the count matrix $X\in\mathbb{R}^{n\times p}$ by pitting
two Stockfish \cite{wiki:stockfish} engines against each other and
logging their moves. In order to investigate how pieces of the same
type are similar to or different from each other, we let each row
of $X$ correspond to a piece rather than a piece type. Hence, we
set $n$ to $16$, which is the number of white pieces.

We let the engines play against each other for $4{,}000{,}000$ turns,
resulting in $23{,}096$ games with an average length of $173$ turns.
White player wins $45\%$ of the time, black player wins $10\%$ of
the time, and a draw occurs $45\%$ of the time. We observe a high
percentage of draws because repeating the same move three times causes
a draw, and the black player frequently uses that strategy. We only
log moves by the white player from the games which the white player
won, yielding a total of $881{,}779$ moves, which we use to construct
the $X\in\mathbb{R}^{16\times4096}$ matrix. We then normalize each
row by its total counts, scale the columns to have zero mean and unit
variance, and apply PCA. 

\begin{figure}
\centering{}\includegraphics[scale=0.45]{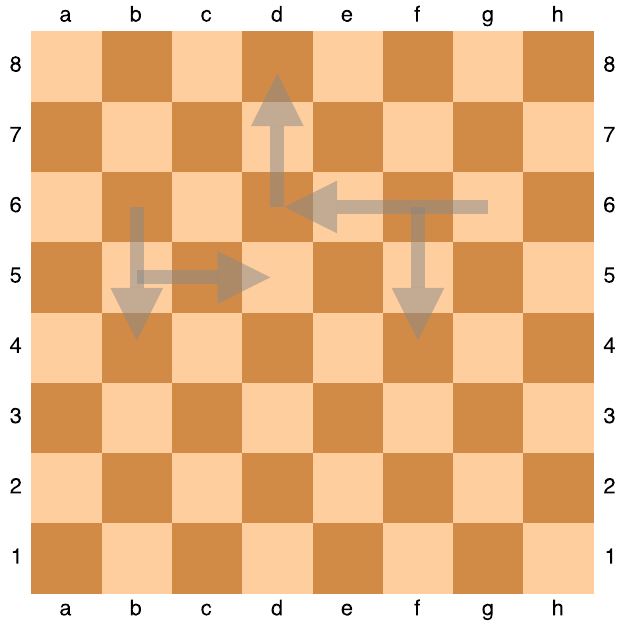}\includegraphics[scale=0.45]{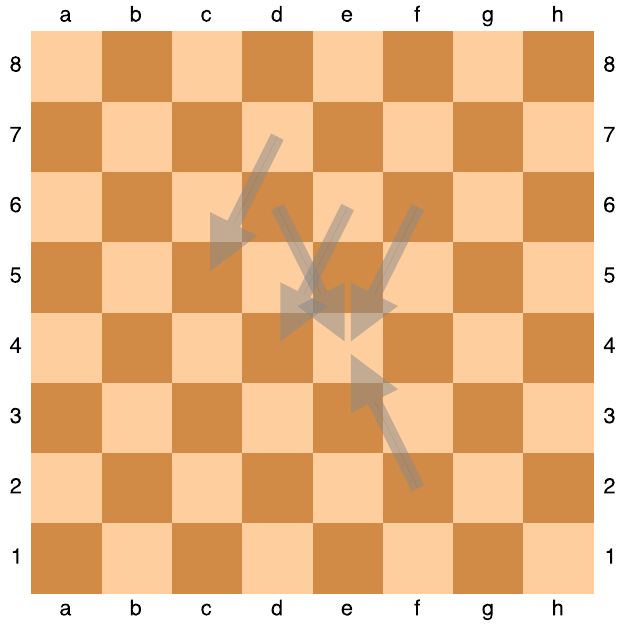}\includegraphics[scale=0.45]{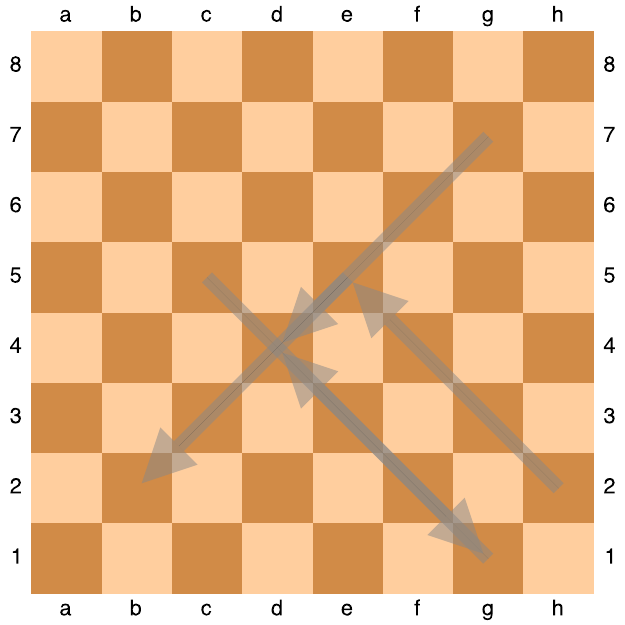}\caption{\label{fig:pca-stockfish-boards}The top five moves associated with
the Stockfish loading vectors.}
\end{figure}

\begin{figure}
\begin{centering}
\includegraphics[scale=0.75]{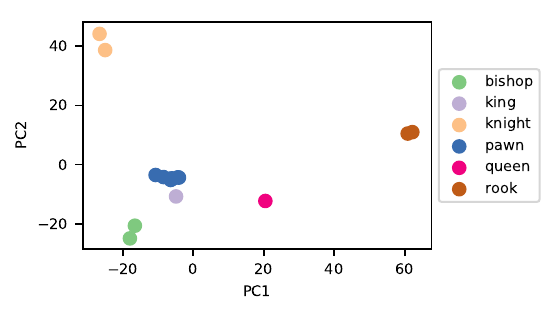}\includegraphics[scale=0.75]{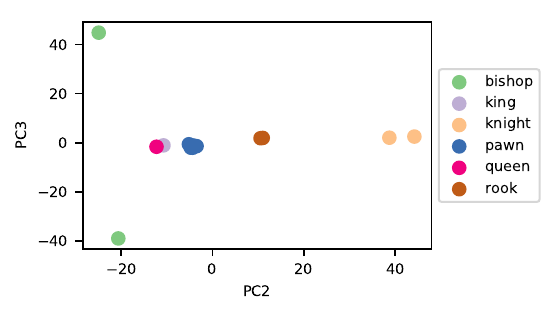}
\par\end{centering}
\begin{centering}
\caption{\label{fig:pca-stockfish}Principal component scores of chess pieces
(Stockfish moves). Left: the first principal component versus the
second. Right: the second principal component versus the third.}
\par\end{centering}
\end{figure}

We visualize the first three components in Figure \ref{fig:pca-stockfish}.
Each loading vector can be interpreted as a fundamental building block
of expert chess moves, and the principal component scores depict how
each piece can be described in terms of these building blocks. The
first five principal components explain $81.0\%$ of the variance
in the movement data, at which point adding more principal components
does not increase the explained variance by much. This is because,
as seen in Figure \ref{fig:pca-stockfish}, pieces of same type have
similar principal component scores, and they share the same loading
vectors to construct their movement patterns. This is surprising,
since PCA has no explicit information that these pieces belong to
the same type, and despite their similarities, individual pieces such
as pawns do have their own unique movement patterns within the $4096$-dimensional
movement space.

In figure \ref{fig:pca-stockfish-boards}, we display the highest
scored moves associated with the loading vectors. The third dimension
is the first dimension in which pieces of the same type, namely bishops,
are not clustered together with respect to their component scores.
In fact, the bishop which moves among black squares has a high component
score, and the bishop which moves among white squares has a low component
score.  

These decompositions are helpful in uncovering fundamental movement
patterns and understanding the relationship between pieces with respect
to these patterns. To the best of our knowledge, this is the first
attempt to use machine learning to conduct such an analysis. In the
process, we have also obtained our first $d$-dimensional vector representations,
namely $W=XC^{\star}$. 

\section{Position-dependent piece vectors}

In this section, we attempt to improve upon the vector representations
we have obtained so far. In particular, one potential issue is that
the piece vectors we constructed are constant with respect to chess
positions. For example, the vector representation of a pawn does not
vary based on the remaining pieces on the board. We propose a hash-based
method to address this issue. We also conduct our first quantitative
evaluation.

\subsection{Zobrist hashing}

Zobrist hashing \cite{Zobrist1970} is a method for constructing a
universal hash function that was originally developed for abstract
board games. It is typically used to build transposition tables which
prevent game agents from analyzing the same position more than once.

A natural way to generate position-dependent piece vectors is to modify
the construction of the count matrix $X\in\mathbb{R}^{n\times p}$.
Previously, each row of $X$ corresponded to a piece or piece type.
This time, we expand the rows to correspond to the Cartesian product
of pieces and hash buckets. The hash buckets partition the space of
all chess positions via Zobrist hashing. Thus, instead of generating
a unique vector representation for each piece, we generate it for
each piece given a hash bucket, where each bucket represents a random
collection of chess positions. In our experiments, we let the number
of hash buckets go up to $32{,}768$. Since there are $16$ pieces,
this yields the sparse count matrix $X\in\mathbb{R}^{524{,}288\times4096}$.

\subsection{Non-negative matrix factorization (NMF)}

In addition to expanding the count matrix, we also switch our matrix
factorization method from PCA to NMF \cite{Lee1999}. NMF is specified
by the optimization problem
\begin{equation}
\begin{aligned} & \underset{W,H}{\text{minimize}} &  & \left\Vert X-WH\right\Vert _{F}^{2}\\
 & \text{subject to} &  & W\geq0,H\geq0
\end{aligned}
.\label{eq:nmf-objective}
\end{equation}
In the NMF objective \eqref{eq:nmf-objective}, $W\in\mathbb{R}^{n\times d}$
encodes the $d$-dimensional vector representations and $H\in\mathbb{R}^{d\times p}$
encodes the $p$-dimensional basis vectors. The corresponding matrices
in the PCA objective \eqref{eq:pca-objective} are $W\simeq XC$ and
$H\simeq C^{T}$, where $\simeq$ simply denotes a semantic (but not
a numeric) similarity. 

We switch from PCA to NMF for a couple of reasons. First, even though
both PCA and NMF attempt to minimize the reconstruction error, the
positivity constraints of NMF are more flexible than the orthonormality
constraints of PCA. In fact, NMF yields lower reconstruction error
than PCA, and as we shall see below, reconstruction error plays an
important role in predicting chess moves. Second, NMF yields a different
set of vectors than PCA which are also worth investigating.

We demonstrate how matrix reconstruction can be used to predict chess
moves from chess positions. As discussed, each row of $W\in\mathbb{R}^{n\times d}$
encodes a $d$-dimensional vector representation, denoted as $w_{i,j}^{T}\in\mathbb{R}^{1\times d}$,
that is associated with a white piece $i\in\left\{ 1,\ldots,16\right\} $
and a hash bucket $j$. Each hash bucket $j$ corresponds to a random
collection of chess positions. Let $w_{0,.}$ be the vector representation
for the special piece $0$ that is always assigned to the zero vector.
Let $s\in\mathcal{S}$ represent a chess position, which is a description
of the locations of all the pieces on a chess board. Let $f_{l}:\mathcal{S}\to\{0,...,16\}$
be a function that maps a chess position to one of the $17$ pieces
that can be located on board location $l\in\{1,...,64\}$, where $f_{l}\left(s\right)=i$
if board location $l$ contains a white piece $i$, and $f_{l}\left(s\right)=0$
if $l$ is empty or contains a black piece. Let $h:\mathcal{S}\to\mathcal{J}$
be the Zobrist hash function that maps a chess position to a hash
bucket. Then, given a chess position $s$, $y\left(s\right)=\sum_{j=1}^{64}H^{T}w_{f_{j}\left(s\right),h\left(s\right)}$
is the $4096$-dimensional prediction vector, and $\argmax_{_{i}}y\left(s\right)_{i}$
is the index of the predicted move.

The intuition behind this prediction is as follows. Each row of $X$,
denoted as $x_{i,j}^{T}\in\mathbb{R}^{1\times4096}$, represents the
number of times a white piece $i$ has been associated with a particular
move for chess positions in $\mathcal{S}_{j}=\left\{ s\in\mathcal{S}\mid h\left(s\right)=j\right\} $.
Since equation \ref{eq:nmf-objective} minimizes reconstruction error,
given a chess position $s$, $H^{T}w_{f_{j}\left(s\right),h\left(s\right)}\approx x_{f_{j}\left(s\right),h\left(s\right)}$.
Thus, $y\left(s\right)\approx\sum_{j=1}^{64}x_{f_{j}\left(s\right),h\left(s\right)}$
approximates the number of times a move has been selected for all
the chess positions that are hashed into bucket $j$. 

\subsection{Quantitative evaluation}

In this subsection, we evaluate the accuracy of making move predictions
using matrix reconstruction and position-dependent piece vectors.
We split the Stockfish data, which consists of $881{,}779$ moves,
into $80\%$ train and $20\%$ test. We use the train data to generate
the count matrix $X\in\mathbb{R}^{524{,}288\times4096}$, apply NMF,
and obtain $W_{\text{train}}$ and $H_{\text{train}}$. We then iterate
over each chess position $s$ in test data, replace their bitboard
vectors with the relevant rows from $W_{\text{train}}$, and compute
$y\left(s\right)$.

\begin{figure}
\begin{centering}
\includegraphics[width=13cm,height=5cm]{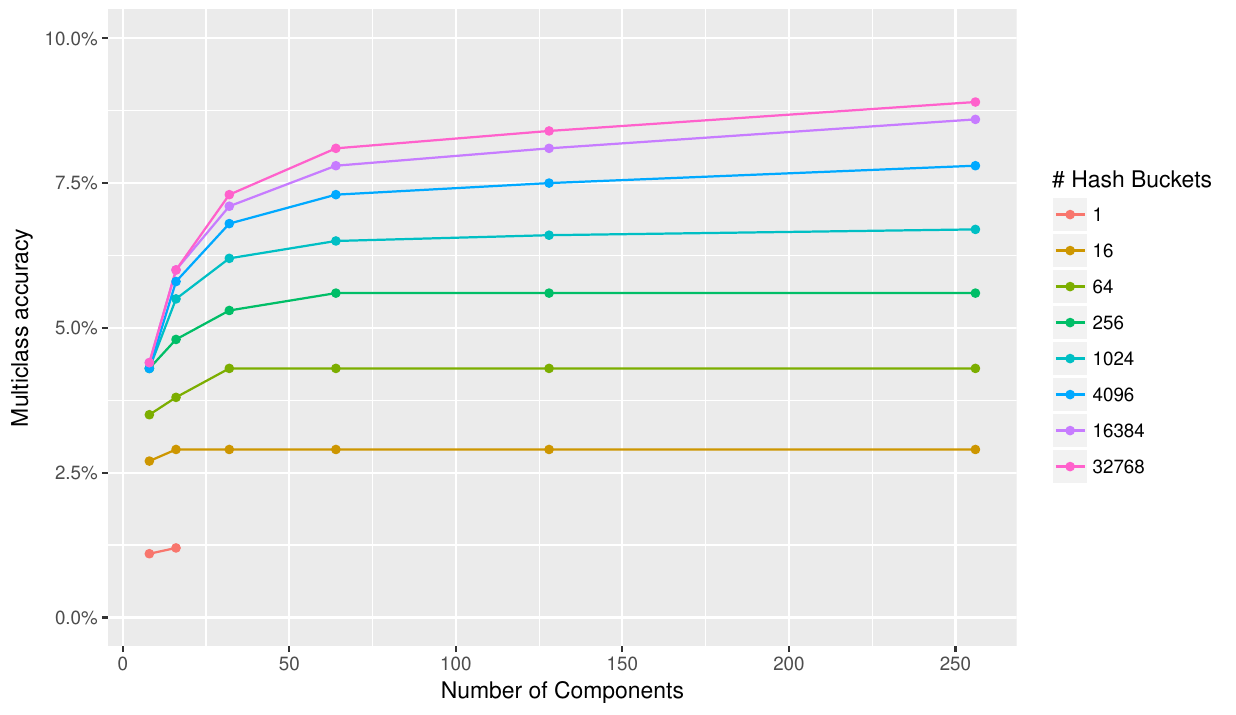}\caption{\label{fig:nmf-test-mca}Test multiclass accuracy of matrix reconstruction
with position-dependent piece vectors. }
\par\end{centering}
\end{figure}

In figure \ref{fig:nmf-test-mca}, we display the multiclass accuracy
on held-out test data. As we increase the number of hash buckets,
the held-out accuracy improves. With $32768$ hash buckets, this method
predicts $8.8\%$ of Stockfish moves correctly, on a task with $4096$
classes.

\bibliographystyle{plain}
\bibliography{chess2vec_short}

\end{document}